# Evaluating the Diagnostic Classification Ability of Multimodal Large Language Models: Insights from the Osteoarthritis Initiative

Li Wang[1,2]; Xi Chen[1,2]; XiangWen Deng[3]; HuaHuiYi[4]; ZeKun Jiang[4]; Kang Li[4]; Jian Li[1,2]

1 Sports Medicine Center, West China Hospital, West Chian School of Medicine, Sichuan University, Chengdu, Sichuan, China

2 Department of Orthopedics and Orthopedic Research Institute, West China Hospital, Sichuan University, Chengdu, Sichuan, China

3 Eller College of Management, The University of Arizona

4 West China Biomedical Big Data Center, West China Hospital, Sichuan University, Chengdu, China

## Abstract

Multimodal large language models (MLLMs) show promising performance on medical visual question answering (VQA) and report generation, but these generation and explanation abilities do not reliably transfer to disease-specific classification. We evaluated MLLM architectures on knee osteoarthritis (OA) radiograph classification, which remains underrepresented in existing medical MLLM benchmarks, even though knee OA affects an estimated 300 to 400 million people worldwide. Through systematic ablation studies manipulating the vision encoder, the connector, and the large language model (LLM) across diverse training strategies, we measured each component's contribution to diagnostic accuracy. In our classification task, a trained vision encoder alone could outperform full MLLM pipelines in classification accuracy and fine-tuning the LLM provided no meaningful improvement over prompt-based guidance. And LoRA fine-tuning on a small, class-balanced dataset (500 images) gave better results than training on a much larger but class-imbalanced set (5,778 images), indicating that data balance and quality can matter more than raw scale for this task. These findings suggest that for domain-specific medical classification, LLMs are more effective as interpreters and report generators rather than as primary classifiers. Therefore, the MLLM architecture appears less suitable for medical image diagnostic classification tasks that demand high certainty. We recommend prioritizing vision encoder optimization and careful dataset curation when developing clinically applicable systems.

# Introduction

Knee osteoarthritis (OA) is one of the leading causes of global disability, affecting 300-400 million people[1] worldwide and significantly contributing to the global disability burden[2,3]. X-ray is considered the gold standard for knee OA screening, and the Kellgren-Lawrence (KL) grading system remains the most widely used method for assessing disease severity[4]. However, accurate interpretation of X-rays relies heavily on the expertise of radiologists and orthopaedic surgeons, often leading to variability in grading and potential delays in diagnosis[4,5]. Given the high prevalence of knee OA, there is a critical need to accurately detect and quantify its severity. Automated grading systems can provide objective and reproducible predictions that are immune to observer fatigue, and artificial intelligence (AI) models have shown promise in automating this process[6,7].

Recently, Multimodal Large Language Models (MLLMs), such as Med-PaLM M and LLaVA-Med, have expanded the scope of medical image analysis from single-modality tasks to multimodal applications, such as visual question answering and radiology report generation. By integrating vision and language alignment, these models enable multimodal understanding and generation capabilities for both images and text, showing potential for end-to-end medical image analysis in diagnostics[8-10]. However, despite their impressive generative capabilities in general medicine, their reliability in disease-specific tasks, such as distinguishing subtle stages of knee OA, remains questionable[9,11,12]. Notably, even in general domains beyond medicine, the classification capabilities of MLLMs are frequently criticized for being inferior to their own vision encoders or specialized classification models[13-16]. Therefore, misclassification or missed diagnoses of common disease-specific labels by MLLMs, together with their intrinsic black-box properties[17] and the hallucination phenomena characteristic of LLMs[18], would substantially constrain the clinical deployment of these models[19,20], despite the enormous data and computational resources invested in their training. There is also a critical need to validate whether adding an LLM actually improves diagnostic accuracy compared to specialized visual analysis alone.

In this study, we addressed these issues by evaluating the performance of MLLM architectures specifically for knee OA classification. We conducted systematic ablation studies using different training strategies to assess the contribution of each component to diagnostic accuracy, using the radiology images from the Osteoarthritis Initiative (OAI), which is currently absent from the MLLM evaluation benchmarks and training datasets.

# Methods

The research framework is shown in **Figure 1**. Because KL grading[4] is not only a classification task but also contains textual descriptive attributes, we employ prompt engineering to transform the KL grading task into a text-generation task that aligns with the capabilities of MLLMs. This reformulated task is used for model training and evaluation (**Figure 1B** and **Figure 1C**). In addition, several widely used general-purpose MLLMs are included as comparative baselines. Approval for data use was obtained through the official OAI website.

**Data**

We selected knee X-ray images from the OAI dataset and we chose the publicly available version provided by Chen et al.[21] with unified image preprocessing and labeling, which has been adopted in several studies[5,22,23]. The processed dataset by Chen et al. consists of 5,778 images in the training set, 826 in the validation set, and 1,656 in the test set, with a ratio of 7:1:2. The detailed distribution of the dataset is shown in **Table 1**, and we constructed corresponding image–text pairs according to the description of each KL grade[4] for subsequent training. The text labels are provided in **Table 2**.

**Training Strategy and Ablation Study**

The open-source LLaVA architecture is currently the commonly used architecture in medical multimodal large models[8,9,24], and the current training paradigm for MLLMs consists of three stages: vision encoder pretraining, connector-only training, and fine-tuning the connector and LLM on downstream tasks. We chose to further train LLaVA-Med, which already possesses extensive medical knowledge, to better adapt the model to our task. Based on the training strategies of LLaVA-Med and LLaVA-Rad[8,9], our training pipeline includes (**Figure 1B**): (1) standalone training of the vision encoder, (2) standalone training of the connector, (3) LoRA fine-tuning with the vision encoder frozen, and (4) various combinations of the above three modes. Since the dataset is class-imbalanced, we constructed two training subsets: one with 100 randomly sampled images per class (500 images total), and another comprising all 5,778 images. Both subsets were used for connector training and LoRA fine-tuning in ablation studies. A detailed description of the trained models and their corresponding training configurations is provided in **Supplementary Table 1.**

For training the vision encoder of LLaVA-Med (clip-vit-large-patch14-336), a model with image–text alignment capability from the CLIP series[25], we constructed corresponding image–text pairs according to the description of each KL grade[4]. The text labels are provided in **Table 2**. We adopted the standard cross-entropy loss as the objective function, with a batch size of 32, a learning rate of 1e-5, a weight decay of 1e-5, and 5 warmup epochs. The maximum number of epochs was set to 120, and the best model and early stopping were determined based on the performance of the confusion matrix.

For the connector (Multilayer Perceptron, MLP) training, we referenced the official LLaVA repository implementation[26], with a batch size of 16, a learning rate of 1e-4, and training for one epoch.

For the LoRA fine-tuning, the prompt template is shown in **Table 2**. The training procedure followed the official LLaVA implementation, with a batch size of 1a6, learning rate of 2e-4, LoRA rank of 128, LoRA alpha of 256, MLP learning rate of 2e-5, a warmup ratio of 0.03 and the model was trained for three epochs in total.

All training was performed on an NVIDIA A800 GPU (80 GB) equipped with an Intel Xeon(R) Gold 6348 processor, and the remaining training details can be found in https://github.com/wanglihx/LLaVA-OA.

**Evaluation and Data Analysis**

In addition to the models trained in this study, several baseline models were included for comparison. Specifically, we evaluated the vision encoder of LLaVA-Med (clip-vit-large-patch14-336), the LLaVA-Med model (llava-med-v1.5-mistral-7b), and four general MLLMs (GPT-4o, GPT-5, Gemini-2.5-Pro and Qwen-VL-Max) as baselines. **Supplementary Table 1** provides details of all the models included in the evaluation.

For the CLIP models, both before and after training, we adopted a zero-shot classification approach. For the MLLMs, evaluations were performed using the same prompt templates as those employed during LoRA fine-tuning and the temperature was all set at 0.01 except GPT-5, which does not support this parameter. In addition, the maximum token limit (max_token) was set to 2048 for GPT-5 and Gemini-2.5-Pro due to their built-in reasoning modes, and 512 was set for all other models. Model performance was analyzed from both binary and multi-class perspectives based on the KL grading system for knee OA, with KL = 2 defined as the diagnostic cutoff point for the binary classification[27].

The evaluation metrics of the classification ability included Accuracy, Precision, Specificity, Sensitivity, and F1-Score[28]. To assess classification agreement between models and between each model and the reference standard, we computed Cohen's Kappa, linearly weighted Cohen's Kappa, and the Matthews Correlation Coefficient (MCC)[28].

For the fine-tuned CLIP models, we additionally plotted Receiver Operating Characteristic (ROC) and Precision–Recall (PR) curves for both binary and multi-class classification tasks and calculated the corresponding Area Under the Curve (AUC) and Average Precision (AP), respectively[28]. Furthermore, we applied Gradient-weighted Class Activation Mapping (Grad-CAM)[5,29] to the CLIP models to visualize the spatial regions contributing to the model's classification decisions, providing explainability of the visual features learned. All data analyses and visualizations were performed using Python (version 3.9.7).

**Data and Code availability**

The code is available at GitHub: https://github.com/wanglihx/LLaVA-OA; the training data are available at https://data.mendeley.com/datasets/56rmx5bjcr/1; the weights of the trained vision encoder are available at https://huggingface.co/wanglihx/CLIP-OA; the above data and code ensure that the study is fully reproducible.

## Results

Overall, based on the training strategy and final evaluation results, we categorized all models into three groups: baseline models, models with trained vision encoder and connector, and models where either vision encoder or connector was not trained. All training curves are presented in **Supplementary Figure 1**. A comprehensive evaluation of all models, including related metrics, as well as the prompts and text labels used during evaluation, is provided in **Supplementary Table 2,** and four models were excluded from evaluation due to non-convergent training curves or outputs consisting entirely of hallucinations (**Supplementary Table 3**).

### Baseline models

From the confusion matrices of both multiclass (**Figure 2**) and binary classification (**Figure 3**), all baseline models can be considered to exhibit random guessing or minimal classification capability for knee OA. In the multiclass task, CLIP, LLaVA-Med, and Qwen-VL-Max concentrated their predictions on one or two specific categories. Gemini-2.5-Pro predicted nearly all images (92.7%) as KL grade 2–4 for knee OA, and marked 9 images as unassessable due to poor image quality. Although GPT-5 and GPT-4o achieved overall multi-class accuracies of only 40.58% and 38.89%, respectively, both models demonstrated a certain level of classification performance on severe OA. GPT-5 correctly classified 66.7% of KL grade 4 cases as KL grade ≥2. GPT-4o classified 56.1% and 80.4% of KL grade 3 and KL grade 4 cases as KL grade ≥2, respectively.

### Trained Vision encoder Evaluation and Visualization Results

The CLIP model as vision encoder exhibited significant changes before and after training. Based on the variations of the confusion matrix of the validation set (**Supplementary Figure 2**), we selected the model from the sixth epoch as the optimal model, as its misclassifications were mainly between adjacent KL grades, indicating that the model had captured the ordinal continuity among grades, unlike earlier or later epochs that showed more cross-grade confusion. The optimal model achieves accuracies ranging from 81.2% to 86.3% across the KL grades except KL grade 1 on the test set (**Figure 2**). The accuracy for KL grade 1 was only 15.2%. CLIP-OA predicted 40.8% and 42.2% of KL grade 1 (doubtful OA) cases as KL grade 0 (normal knee) or KL grade 2 (mild OA), respectively, with average confidences of 0.838 and 0.795 (Supplementary Figure 3). In contrast, the 15.2% of cases correctly predicted as KL grade 1 exhibited a relatively low average confidence of only 0.582. Consider that this result aligns with the CLIP model's semantic understanding of the descriptors (i.e., normal, doubtful, and mild), with the model tending toward making more definitive judgments from its perspective. In terms of overall multi-class classification, CLIP-OA was comparable to dedicated classification models on the same test set (**Supplementary Table 4**), and achieved the best accuracy for KL grade 2-4. For the binary classification task, CLIP-OA demonstrated reliable performance, with an AP of the PR curve of 0.945 (95% CI: 0.933, 0.955) and AUC of the ROC curve of 0.949 (95% CI: 0.939, 0.958) (**Figure 4**), outperforming binary change detection models trained on larger sample sizes using Siamese neural networks[27] (PR AP 0.75-0.81, ROC AUC 0.85-0.90).

Grad-CAM visualization (**Figure 5**) showed changes in CLIP attention across different KL grades before and after training. Heatmaps of CLIP and CLIP-OA were mutually corroborative with the classification results, demonstrating a clear distinction between models that "understand" versus "do not understand" the radiographic features. CLIP-OA consistently focused attention on the medial and lateral joint spaces and marginal osteophytes across all KL grades. In contrast, CLIP failed to identify salient features and could only make random guesses between KL grade 1 and KL grade 2 for all images. These visualization results further suggest that the vision encoder serves as the core component for task-specific classification in MLLMs.

**Models with Trained Vision Encoder and Connector**

The five models ( LLaVA-Med-CLIPOA-MLP500, LLaVA-Med-CLIPOA-MLP5778, LLaVA-Med-CLIPOA-MLP500-LoRA500, LLaVA-Med-CLIPOA-MLP5778-LoRA500, LLaVA-Med-CLIPOA-LoRA500) that trained the vision encoder and connector demonstrated relatively high inter-model agreement (linearly weighted κ = 0.894–0.943), surpassing the agreement observed between each model and its respective vision encoder (linearly weighted κ = 0.777–0.846) and MCC showed a similar pattern (**Figure 6**). Compared to their own vision encoder CLIP-OA, these five models showed improvements in KL grade 1 accuracy (50.3% to 62.8% versus 15.2%) but declines in KL grade 0 (64.5% to 76.1% versus 82.0%) and KL grade 2 (49.2% to 67.3% versus 81.2%). Correspondingly, in the binary classification task, these five models exhibited lower sensitivity (70.7% to 81.7% versus 92.9%) and higher specificity (92.0% to 97.2% versus 79.4%) compared to CLIP-OA. Since MLLMs cannot output confidence scores due to their generative nature, they are unable to generate PR and ROC curves as accurately as CLIP-OA.

When using the full training dataset of 5,778 images, which exhibits severe class imbalance, the four LoRA fine-tuned LLM models experienced training curves that failed to converge (**Supplementary Figure 1c, 1d ,1m**), or when the training curve converged, severe hallucinations occurred during responses (**Supplementary Table 3**). In contrast, a vision encoder trained independently does not appear to be substantially affected by class imbalance. CLIP-OA achieves an accuracy of 86.3% on the minority class KL grade 4 without any specific techniques for handling class imbalance.

**Models without Trained Vision encoder or Connector**

For LLaVA-Med-CLIPOA, we directly replaced the initialized vision encoder with a pre-trained one without further training the MLP and LLM, yet it still failed to perform classification (**Figure 2**). For LLaVA-Med-MLP5778-LoRA500 and LLaVA-Med-MLP500-LoRA500, when training only the MLP and LLM without training the vision encoder, we found that they also completely failed to perform classification **(Figure 2)**. In other words, following LLaVA-Med's training strategy cannot succeed on this task because it does not train the vision encoder. Moreover, even if the vision encoder is trained but the connector is not trained for further alignment, it also cannot succeed.

## Discussion

Our study demonstrates that the diagnostic reliability of MLLMs for Knee OA is fundamentally dependent on the model's ability to accurately identify radiographic features. However, the generative nature of LLMs does not enhance this process and may even introduce uncertainty, as our agreement analysis reveals that the final LLM output frequently alters the classification results provided by the vision encoders. In contrast to previous studies that merely evaluated overall capabilities, specifically two previous studies[11,12] that assessed GPT-4o's KL grading on limited sample sizes (200 and 117 images) and noted its poor multiclass performance, our research offers an in-depth exploration into why MLLMs underperform in this domain. Furthermore, given that standalone vision encoders are capable of achieving strong classification performance, we critically investigate the incremental value of integrating an LLM in this context.

Furthermore, given the current absence of established LLM-integrated AI models in the field of rheumatic diseases, our study offers a roadmap for training and developing clinically viable image analysis models. Our results support a framework where the LLM component is ideally reserved for downstream tasks, such as summarizing patient history or generating patient-friendly explanations, rather than executing the primary image diagnosis. This functional separation ensures that the core diagnostic task aligns with the medical standard of care. Notably, this approach has already been validated in ophthalmology, where Wu et al.[30] conducted a three-arm randomized controlled trial in near-clinical settings to compare the performance of clinicians alone, clinicians assisted by AI, and AI alone. Interestingly, in their methodological description for image tasks, the study explicitly states: "For downstream image classification tasks, we keep only the encoder from the image module pretraining step and discard the decoder[30]." In other words, the LLM is directly removed for this task and replaced with an MLP as a classification head to specifically perform image classification. The classification results, along with other textual modalities such as patient history, then serve as input to the LLM to complete downstream tasks like diagnosis and treatment recommendations. This further demonstrates that in medical settings, the most critical component for image analysis is the vision encoder-centered approach for performing explicit classification tasks. Meanwhile, our study on the specific task of knee OA using different training strategies also reveals the central role of the vision encoder.

Of course, we do not deny that MLLM architectures have shown promising results in certain specialized domains. For instance, PathChat[24], which builds upon a foundational vision encoder[31] for pathology and follows the two-stage training strategy of LLaVA-Med, was trained on 456,916 instructions with 999,202 question–answer turns. Compared with LLaVA-Med, it achieved a 63.8% improvement on multiple-choice diagnostic tasks, which can be regarded as a multi-class classification problem. However, this gain likely stems mainly from the foundational vision encoder for pathology rather than the large number of instruction–answer pairs. In other words, the training of the vision encoder is the core factor that endows MLLMs with diagnostic classification ability. Combined with our findings, this further explains why MLLMs that have undergone vision encoder training[9,10,24] tend to substantially outperform LLaVA-Med, which keeps its vision encoder frozen.

The failure of general MLLMs such as GPT-5 on our task is to some extent inevitable, as we believe their vision encoders have not been trained on our specific task, rather than their LLMs lacking knowledge of knee OA. For training beyond the vision encoder, training the connector primarily aligns image and text semantics, effectively enabling the LLM to function as a classification component for medical tasks. Further fine-tuning of the LLM mainly allows it to transform various forms of open-ended questions into the corresponding classification tasks.

Moreover, medical MLLMs often employ large-scale image-text pairs when fine-tuning the LLM component, which demands substantial computational resources and consumes more computational resources than fine-tuning LLMs with text alone. In contrast, our study completes the training of the vision encoder and connector without fine-tuning the LLM, and then guides the MLLM to accomplish specific classification tasks through prompt engineering, thereby achieving a particular response style. Regarding LLM fine-tuning, according to LIMA's Superficial Alignment Hypothesis and related studies[32], pre-trained language models already possess the necessary knowledge, and the role of fine-tuning is to directly update model parameters to guide the model toward a specific and stable response style rather than learning new knowledge[33]. Therefore, using only a small amount of carefully selected data can achieve better results than large-scale data fine-tuning[34]. Our findings suggest that for a specific multimodal medical task, the strategy should be progressive: from prompting to few-shot fine-tuning to full-data fine-tuning. Meanwhile, data distribution should also be considered. In our study, using the full dataset conversely led to training non-convergence and hallucinations, which raises concerns about class imbalance issues. For larger and broader datasets, data distribution and selection need to be carefully considered. For instance, in MIMIC-CXR, fractures and lung lesions account for only 1.7% and 2.7% respectively. Despite being one of the training datasets for LLaVA-Rad, the model ultimately achieved only 18.79% and 14.01% sensitivity for these conditions, respectively. Therefore, it is necessary to pay greater attention to optimizing data selection.

Therefore, regarding the future development of LLM-integrated multimodal AI models in the field of rheumatic diseases like knee OA, our study suggests that, in terms of diagnostic accuracy, explainability, and cost-effectiveness, current MLLM architectures are not well-suited. Although our study has limitations, including the absence of novel algorithms and the use of a single dataset (knee OA X-rays) with two downstream tasks (binary classification for screening and multi-class classification), findings of our study, together with existing research, call for reflection on the suitability and cost-effectiveness of current MLLM architectures for broader medical applications.

**Table 1: KL Grade Definitions and Dataset Distribution**

| Category | Category Description | Number (Training Set) | Number (Validation Set) | Number (Test Set) |
|---|---|---|---|---|
| KL grade 0 | No osteoarthritis, No radiographic features of osteoarthritis. | 2286 | 328 | 639 |
| KL grade 1 | Doubtful osteoarthritis, Doubtful narrowing of joint space and possible osteophytic lipping. | 1046 | 153 | 296 |
| KL grade 2 | Mild osteoarthritis, Definite osteophytes and possible narrowing of joint space. | 1516 | 212 | 447 |
| KL grade 3 | Moderate osteoarthritis, Multiple osteophytes, definite narrowing of joint space, some sclerosis, and possible deformity of bone ends. | 757 | 106 | 223 |
| KL grade 4 | Severe osteoarthritis, Large osteophytes, marked narrowing of joint space, severe sclerosis, and definite deformity of bone ends. | 173 | 27 | 51 |
| Total | | 5778 | 826 | 1656 |

Note. — The Category Description adopted the original wording of the KL grading system[4], which also corresponds to the semiquantitative X-ray assessment items used in the Osteoarthritis Initiative (https://nda.nih.gov/oai).

**Table 2: Text labels for Vision Encoders and the Prompt for MLLMs**

| Text labels for CLIP and CLIP-OA | text_labels = [<br><br>"No osteoarthritis, No radiographic features of osteoarthritis.",<br><br>"Doubtful osteoarthritis, Doubtful narrowing of joint space and possible osteophytic lipping.",<br><br>"Mild osteoarthritis, Definite osteophytes and possible narrowing of joint space.",<br><br>"Moderate osteoarthritis, Multiple osteophytes, definite narrowing of joint space, some sclerosis, and possible deformity of bone ends.",<br><br>"Severe osteoarthritis, Large osteophytes, marked narrowing of joint space, severe sclerosis, and definite deformity of bone ends."<br><br>] |
|---|---|
| Prompt for all MLLMs | You are a professional radiologist. You are provided with a knee X-ray image and you should determine the Kellgren-Lawrence grade of this knee joint X-ray based on the Kellgren-Lawrence grading system.<br><br>The specific criteria for Kellgren-Lawrence grading system are as follows: Grade 0: No osteoarthritis, No radiographic features of osteoarthritis. Grade 1: Doubtful osteoarthritis, Doubtful narrowing of joint space and possible osteophytic lipping. Grade 2: Mild osteoarthritis, Definite osteophytes and possible narrowing of joint space. Grade 3: Moderate osteoarthritis, Multiple osteophytes, definite narrowing of joint space, some sclerosis, and possible deformity of bone ends. Grade 4: Severe osteoarthritis, Large osteophytes, marked narrowing of joint space, severe sclerosis, and definite deformity of bone ends.<br><br>Please output the most likely Kellgren-Lawrence grade you determine. The format of your answer should be: The most likely Kellgren-Lawrence grade of this knee X-ray image is Grade {X}: {the description} |

Note. — For the CLIP-series models, the zero-shot classification implementation essentially requires the model to select, from the five labels above, the text description that best matches the given image. For the MLLMs, the same prompt template was used for LoRA fine-tuning and for testing.

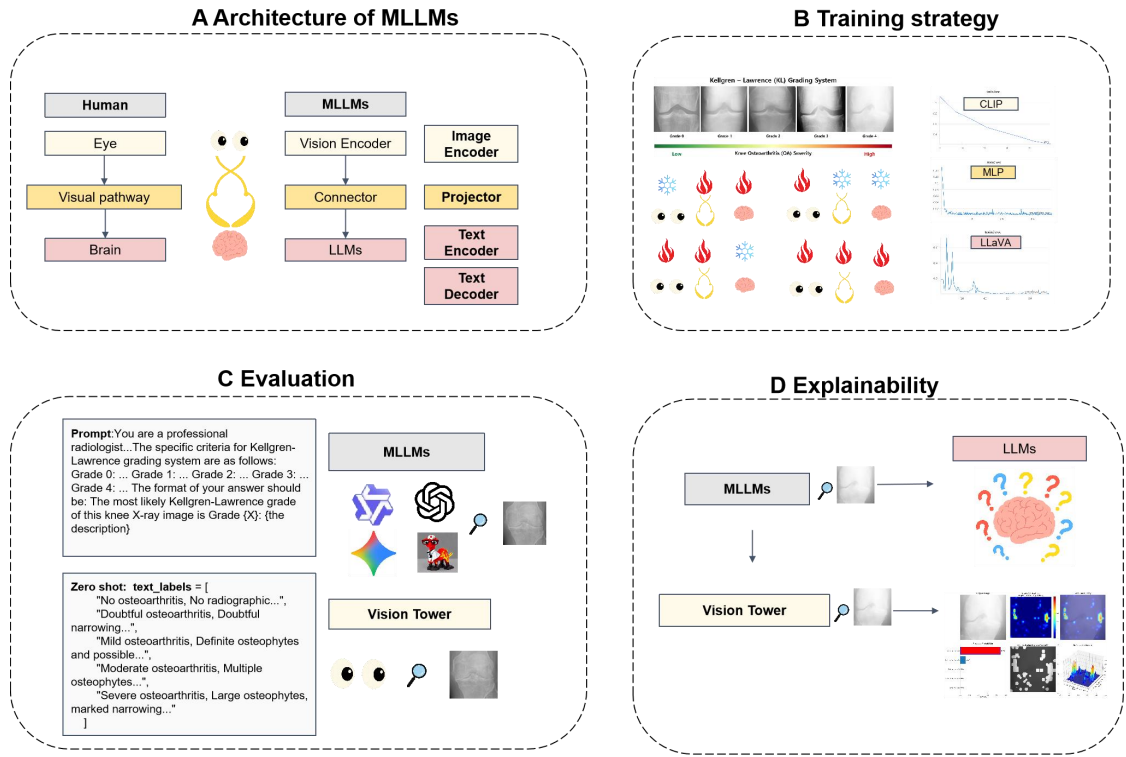

**Figure 1:** Research Framework. **(A)** The architecture of MLLMs. The vision encoder encodes image information, which is projected into the semantic space through a connector and, together with the input prompt, is processed by the text encoder of the LLMs. The final output is generated by the text decoder. **(B)** Different combinations of frozen and trainable modules are explored to identify components critical for domain-specific disease classification. **(C)** MLLMs are evaluated using a unified prompt template, while the vision encoder is assessed using unified text labels for zero-shot classification. **(D)** For classification tasks, the explainability of MLLMs mainly comes from the vision encoder, which can be analyzed using Grad-CAM for visual attention and confidence scores for quantitative assessment, whereas the reasoning process of the LLM component cannot be interpreted in such a precise way.

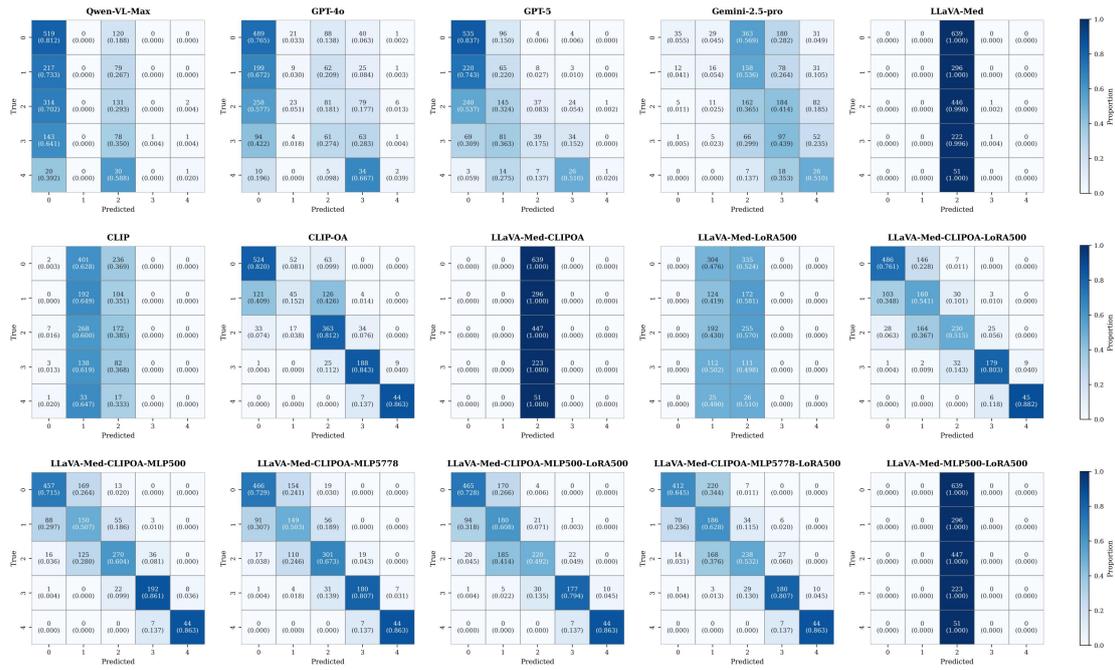

**Figure 2:** Confusion matrices of the baseline model and the MLLMs that can properly follow evaluation prompts in the multi-class classification task. From the diagonal of each confusion matrix, the models that demonstrate the ability to perform KL grade classification include CLIP-OA, LLaVA-Med-CLIPOA-MLP500, LLaVA-Med-CLIPOA-MLP5778, LLaVA-Med-CLIPOA-MLP500-LoRA500, LLaVA-Med-CLIPOA-MLP5778-LoRA500, and LLaVA-Med-CLIPOA-LoRA500.

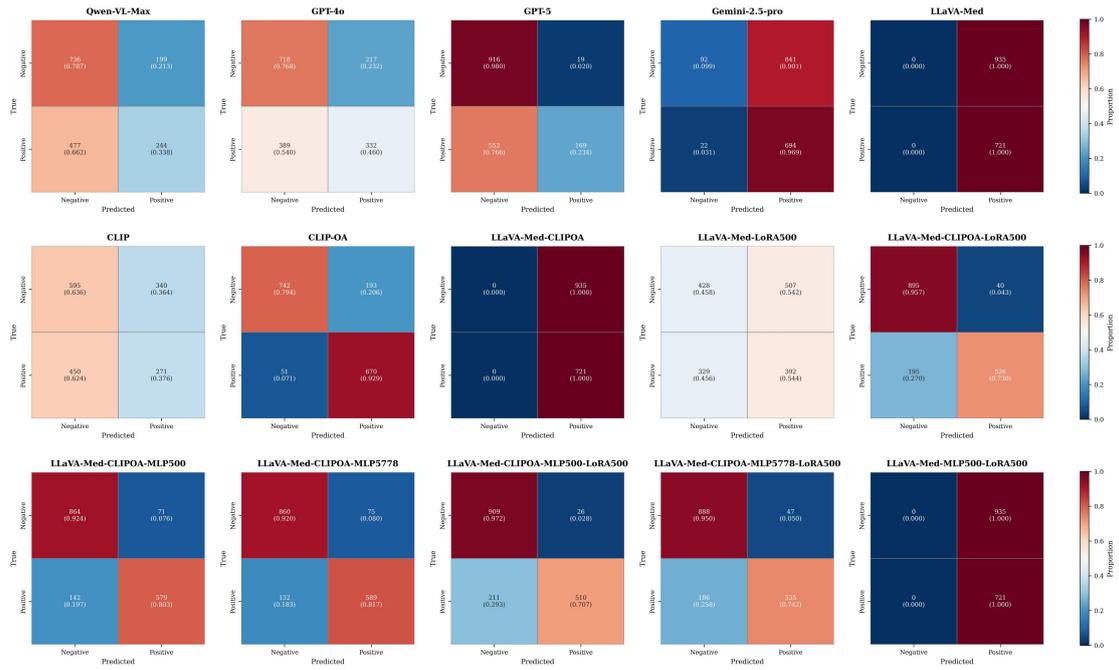

**Figure 3:** Confusion matrices of the baseline model and the MLLMs that can properly follow evaluation prompts in the binary classification task. Using KL grade 2 as the cutoff, the results are aligned with those of the multiclass classification.

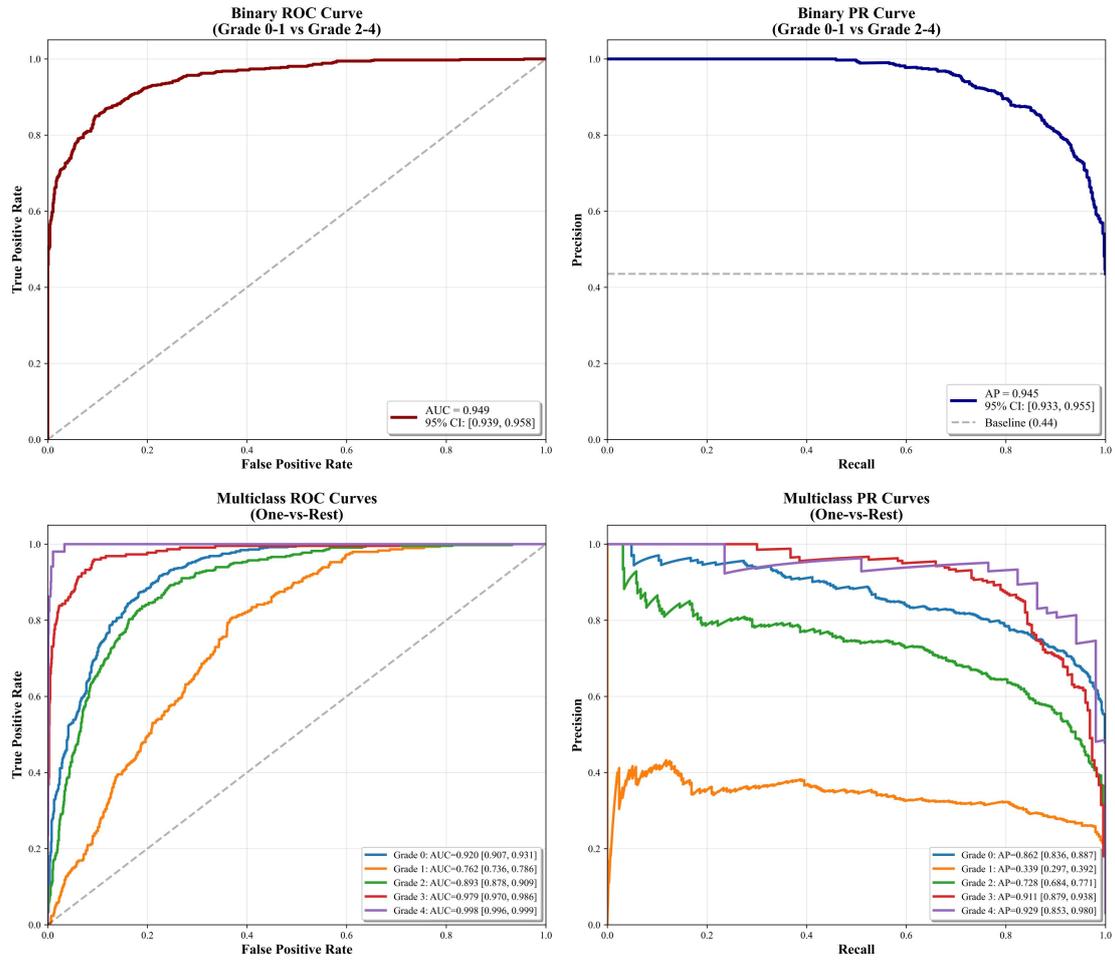

**Figure 4:** ROC and PR curves with corresponding AUC and AP scores for the fine-tuned vision encoder (CLIP-OA) on binary and multi-class classification tasks. The 95% CI was calculated using the Bootstrap method with 2,000 resamples.

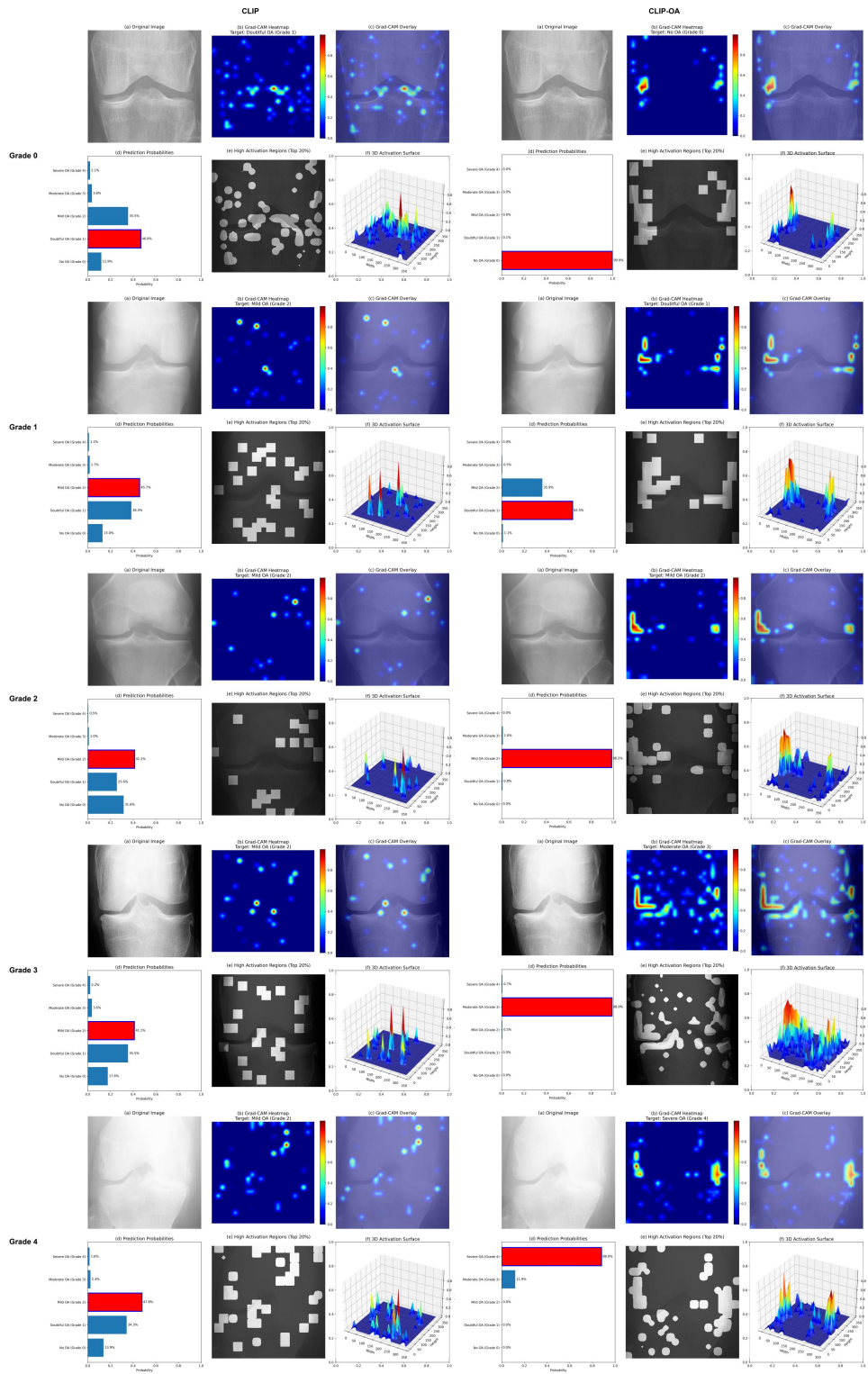

**Figure 5:** Grad-CAM visualizations of the vision encoder before (CLIP) and after training (CLIP-OA) across different KL grades. (a) Original image; (b) Grad-CAM heatmap; (c) Grad-CAM overlay; (d) Confidence output; (e) Top 20% activated regions; (f) 3D activation map.

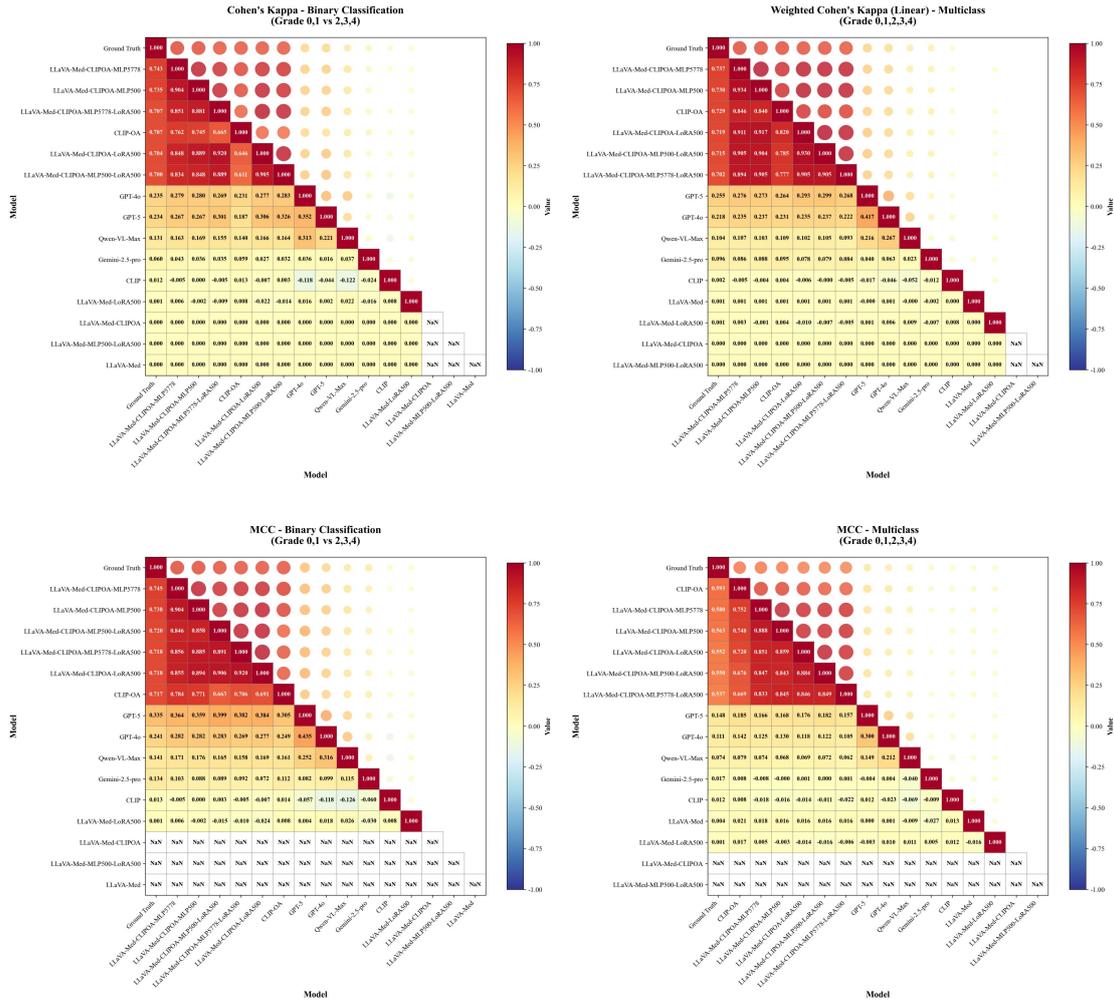

**Figure 6:** Correlation matrices of MCC and Cohen's kappa between the 15 evaluated models and the ground truth for binary and multi-class classification tasks. Cohen's kappa and Matthews Correlation Coefficient (MCC) exhibited consistent trends in our study, although previous research has suggested that MCC is more suitable than Cohen's kappa for class-imbalanced data[35]. This allows us to further compare the classification agreement with that among human experts[4] measured by Cohen's kappa (**Supplementary Table 5**).

**List of Supplementary Materials:**

**Supplementary Figures 1-3**

**Supplementary Tables 1-5**

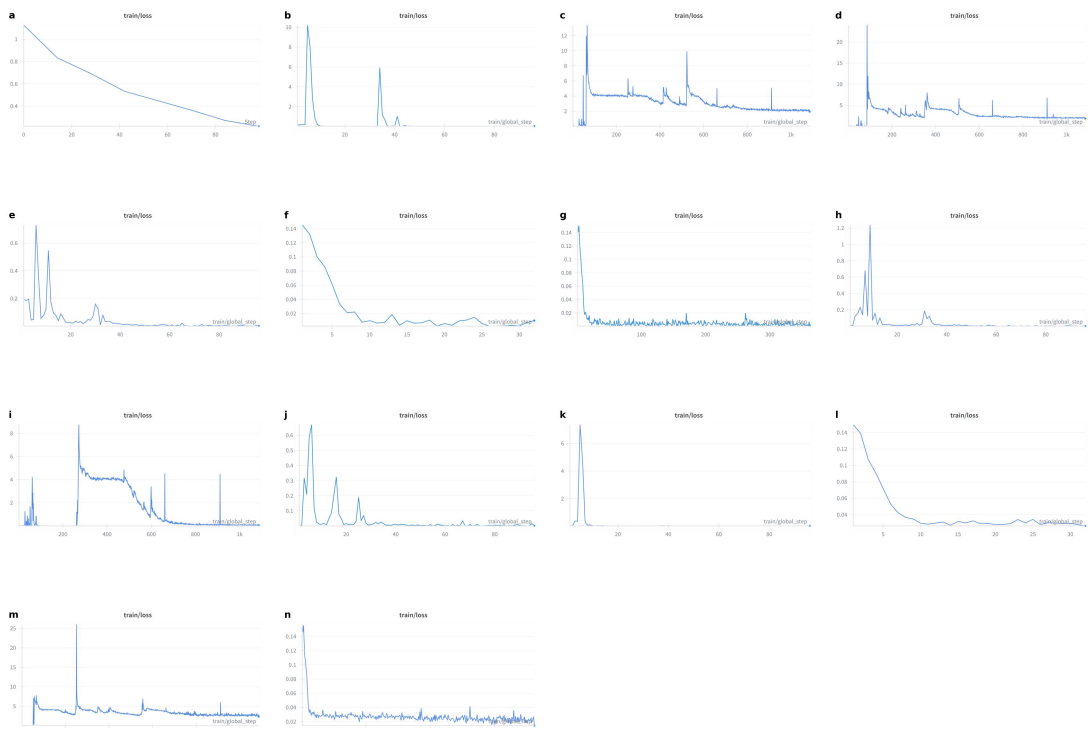

**Supplementary Figure 1:** Training curves. (a) training curve of CLIP-OA

(b) LoRA training curve of LLaVA-Med-LoRA500

(c) LoRA training curve of LLaVA-Med-LoRA5778

(d) LoRA training curve of LLaVA-Med-CLIPOA-LoRA5778

(e) LoRA training curve of LLaVA-Med-CLIPOA-LoRA500

(f) MLP training curve of LLaVA-Med-CLIPOA-MLP500

(g) MLP training curve of LLaVA-Med-CLIPOA-MLP5778

(h) LoRA training curve of LLaVA-Med-CLIPOA-MLP500-LoRA500

(i) LoRA training curve of LLaVA-Med-CLIPOA-MLP5778-LoRA5778

(j) LoRA training curve of LLaVA-Med-CLIPOA-MLP5778-LoRA5778

(k) LoRA training curve of LLaVA-Med-MLP500-LoRA500

(l) MLP training curve of LLaVA-Med-MLP500-LoRA500

(m) LoRA training curve of LLaVA-Med-MLP5778-LoRA5778

(n) MLP training curve of LLaVA-Med-MLP5778-LoRA5778

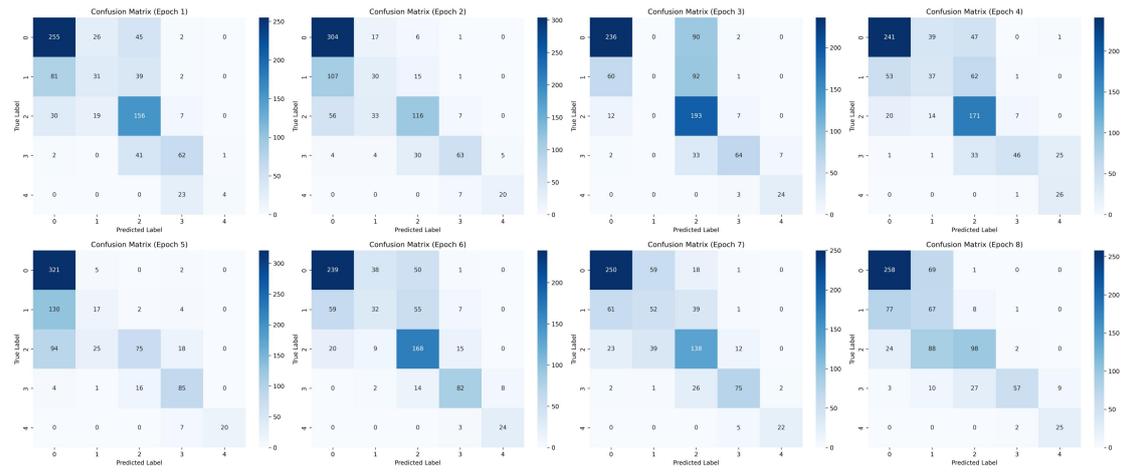

**Supplementary Figure 2:** Confusion matrix evolution on the validation set during the first eight epochs of CLIP-OA training.

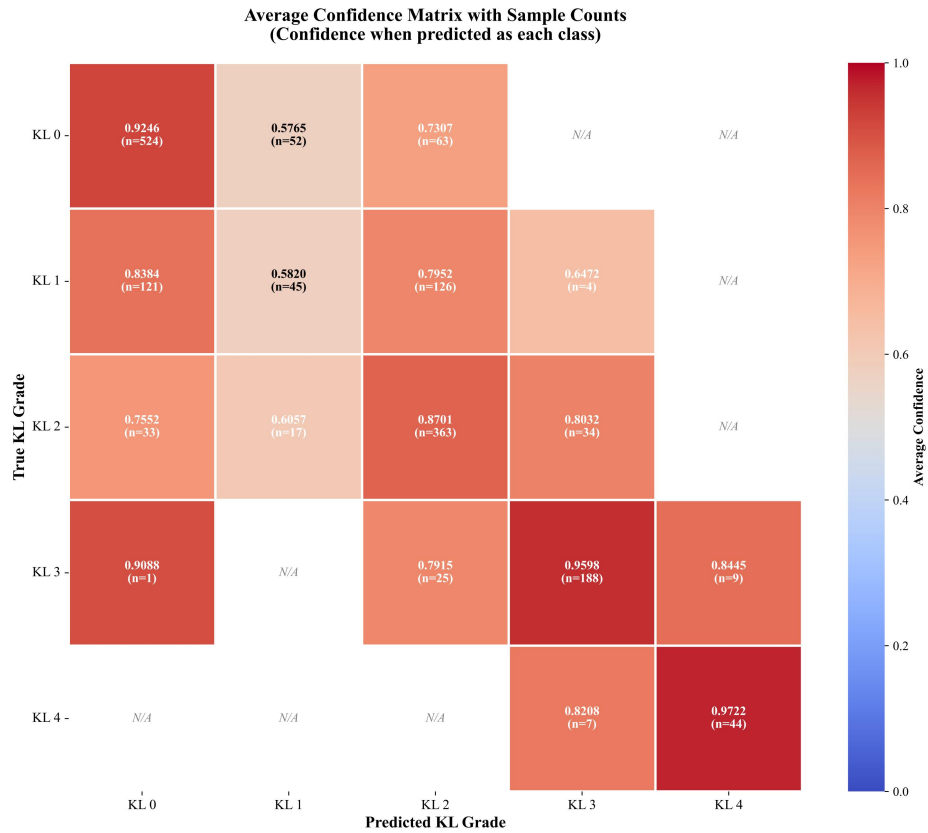

**Supplementary Figure 3:** Average confidence of CLIP-OA predictions on the test set. For example, when the model correctly predicts images with the true label KL-1 as KL-1, the mean confidence is 0.582; when misclassified as KL-0, the mean confidence is 0.838; and when misclassified as KL-2, the mean confidence is 0.795.

**Supplementary Table 1: Details of Baseline Models and the MLLM Variants in the Study**

| Model name | Model details |
| --- | --- |
| GPT-5 | gpt-5-2025-08-07 |
| GPT-4o | gpt-4o-2024-08-06 |
| Qwen-VL-Max | qwen-vl-max-2025-08-13 |
| Gemini-2.5-Pro | gemini-2.5-Pro-2025-06 |
| CLIP | clip-vit-large-patch14-336, the original vision encoder of LLaVA-Med |
| CLIP-OA | Trained CLIP with the dataset of 5778 images |
| LLaVA-Med | llava-med-v1.5-mistral-7b |
| LLaVA-Med-CLIPOA | LLaVA-Med with its vision encoder replaced by CLIP-OA |
| LLaVA-Med-LoRA500 | LoRA fine-tuned LLaVA-Med on a dataset of 500 images |
| LLaVA-Med-LoRA5778 | LoRA fine-tuned LLaVA-Med on a dataset of 5778 images |
| LLaVA-Med-CLIPOA-LoRA5778 | LoRA fine-tuned LLaVA-Med-CLIPOA on a dataset of 5778 images |
| LLaVA-Med-CLIPOA-LoRA500 | LoRA fine-tuned LLaVA-Med-CLIPOA on a dataset of 500 images |
| LLaVA-Med-CLIPOA-MLP500 | Trained the connector (MLP) of LLaVA-Med-CLIPOA separately on a dataset of 500 images |
| LLaVA-Med-CLIPOA-MLP5778 | Trained the connector (MLP) of LLaVA-Med-CLIPOA separately on a dataset of 5778 images |
| LLaVA-Med-CLIPOA-MLP500-LoRA500 | LoRA fine-tuned LLaVA-Med-CLIPOA-MLP500 on a dataset of 500 images |
| LLaVA-Med-CLIPOA-MLP5778-LoRA5778 | LoRA fine-tuned LLaVA-Med-CLIPOA-MLP5778 on a dataset of 5778 images |
| LLaVA-Med-CLIPOA-MLP5778-LoRA500 | LoRA fine-tuned LLaVA-Med-CLIPOA-MLP5778 on a dataset of 500 images |
| LLaVA-Med-MLP500-LoRA500 | Keeping the vision encoder frozen, trained the connector (MLP) alone with 500 images and LoRA fine-tuned with 5778 images. |
| LLaVA-Med-MLP5778-LoRA5778 | Keeping the vision encoder frozen, trained the connector (MLP) alone with 5778 images and LoRA fine-tuned with 5778 images. |

Note. —Four general-purpose models, as well as the LLaVA-Med and CLIP variants, were obtained from their respective official websites or code repositories: GPT-5 and GPT-4o from the OpenAI API (https://openai.com/api/); Qwen-VL-Max from Alibaba Cloud (https://bailian.console.aliyun.com/?admin=1#/home); LLaVA-Med from Hugging Face (https://huggingface.co/microsoft/llava-med-v1.5-mistral-7b); Gemini-2.5-Pro from Google's Gemini API documentation (https://ai.google.dev/gemini-api/docs); and CLIP from Hugging Face (https://huggingface.co/openai/clip-vit-large-patch14-336).

**Supplementary Table 2: Evaluation of MLLMs Using Binary and Multiclass Diagnostic Metrics**

| Model | Binary | | | | | Multiclass | | | |
|---|---|---|---|---|---|---|---|---|---|
| | Accuracy | Precision | Recall/Sensitivity | Specificity | F1 | Accuracy/F1_micro | F1_macro | F1_weighted | MAE±std |
| Qwen-VL-Max | 59.18 | 55.08 | 33.84 | 78.72 | 41.92 | 39.37 | 18.04 | 29.85 | 1.097±1.058 |
| GPT-4o | 63.41 | 60.47 | 46.05 | 76.79 | 52.28 | 38.89 | 23.68 | 32.99 | 1.027±1.004 |
| GPT-5 | 65.52 | 89.89 | 23.44 | 97.97 | 37.18 | 40.58 | 24.09 | 34.25 | 0.908±0.909 |
| Gemini-2.5-Pro | 47.67 | 45.21 | 96.93 | 9.86 | 61.66 | 20.38 | 18.02 | 16.73 | 1.437±1.012 |
| LLaVA-Med | 43.54 | 43.54 | 100.00 | 0.00 | 60.66 | 26.99 | 8.67 | 11.58 | 1.147±0.816 |
| CLIP | 52.29 | 44.35 | 37.59 | 63.64 | 40.69 | 22.10 | 12.41 | 14.18 | 1.065±0.741 |
| CLIP-OA | 85.27 | 77.64 | 92.93 | 79.36 | 84.60 | 70.29 | 67.89 | 67.45 | 0.359±0.595 |
| LLaVA-Med-CLIPOA | 43.54 | 43.54 | 100.00 | 0.00 | 60.66 | 26.99 | 8.50 | 11.47 | 1.147±0.816 |
| LLaVA-Med-LoRA500 | 49.52 | 43.60 | 54.37 | 45.78 | 48.40 | 22.89 | 12.29 | 14.44 | 1.087±0.753 |
| LLaVA-Med-CLIPOA-LoRA500 | 85.81 | 92.93 | 72.95 | 95.72 | 81.74 | 66.43 | 69.70 | 67.63 | 0.361±0.532 |
| LLaVA-Med-CLIPOA-MLP500 | 87.14 | 89.08 | 80.31 | 92.41 | 84.46 | 67.21 | 70.46 | 68.52 | 0.348±0.519 |
| LLaVA-Med-CLIPOA-MLP5778 | 87.50 | 88.70 | 81.69 | 91.98 | 85.05 | 68.84 | 71.85 | 70.08 | 0.337±0.525 |
| LLaVA-Med-CLIPOA-MLP500-LoRA500 | 85.69 | 95.15 | 70.74 | 97.22 | 81.15 | 65.58 | 69.29 | 67.25 | 0.364±0.521 |
| LLaVA-Med-CLIPOA-MLP5778-LoRA500 | 85.93 | 91.92 | 74.20 | 94.97 | 82.12 | 64.01 | 68.65 | 66.15 | 0.379±0.525 |
| LLaVA-Med-MLP500-LoRA500 | 43.54 | 43.54 | 100.00 | 0.00 | 60.66 | 26.99 | 8.50 | 11.47 | 1.147±0.816 |

Note. — MAE, Mean Absolute Error, quantifies the average difference between the predicted and ground-truth KL grades. It is computed as $\frac{1}{N}\sum_{i=1}^{N}|y_i - \hat{y}_i|$, where $N$ is the total number of samples, $y_i$ is the true KL grade (0–4), and $\hat{y}_i$ is the predicted grad

**Supplementary Table 3:** Hallucination Examples.

| Model name | Hallucination Examples | Whether the Loss Converges |
| --- | --- | --- |
| LLaVA-Med-LoRA5778 | likely most most most most most most this this this this this this this this this this this this this de de de de ritritrit. | No |
| LLaVA-Med-CLIP-LoRA5778 | of of of of of of of of of of of of of of of of of of of of of of bone The The The The The The The The ofLLLL of of of of possible possible possible possible... | No |
| LLaVA-Med-CLIPOA-MLP5778-LoRA5778 | ends K. The most likely K, The most of The X spaceell likely K. TheL- The spaceellosisgren- X mostrence of and. The X Xisaw likely K. | Yes |
| LLaVA-Med-MLP5778-LoRA5778 | likely likely likelyell grade gradegren grade grade grade of----rayrayrayrayray :: No o o o o o o o o ostestestestesteste,,.. | No |

Note. — All model responses can be found at https://github.com/wanglihx/LLaVA-OA

**Supplementary Table 4:** Comparison with other studies.

| Author/Model name | Accuracy-KL0 | Accuracy-KL1 | Accuracy-KL2 | Accuracy-KL3 | Accuracy-KL4 | Overall |
|---|---|---|---|---|---|---|
| Chen et al.[1] | 79.2 | 38.5 | 69.8 | 80.7 | 82.4 | 69.7 |
| Yong et al.[2] | 80.4 | 38.5 | 70.0 | 79.8 | 86.3* | 70.2 |
| Feng et al.[3] | 92 | 15 | 70 | 82 | 84 | 70.2 |
| Jain et al.[4] | 82.3 | 38.2 | 73.2 | 80.3 | 84.3 | 71.7 |
| Pi et al.[5] | 89.8* | 39.5* | 79.2 | 83.4 | 84.3 | 76.9* |
| CLIP-OA | 82.0 | 15.2 | 81.2* | 84.3* | 86.3* | 70.3 |

Note. — * The best accuracy.

**Supplementary Table 5:** Reproducibility of the human readers in the study by D. Schiphof et al

| Description of the KL grade | Sample size | Weighted κ value | 95% CI | KL grade ≥1 | | KL grade ≥2 | |
|---|---|---|---|---|---|---|---|
| | | | | % Agreement | κ Value | % Agreement | κ Value |
| Original description | 659 | 0.41 | 0.32 to 0.49 | 76 | 0.32 | 95 | 0.62 |
| Alternative 1 | 144 | 0.66 | 0.53 to 0.79 | 86 | 0.63 | 96 | 0.73 |
| Alternative 2 | 114 | 0.69 | 0.58 to 0.80 | 86 | 0.66 | 94 | 0.72 |
| Alternative 3 | 152 | 0.63 | 0.50 to 0.76 | 83 | 0.53 | 97 | 0.82 |
| Alternative 4 | 158 | 0.35 | 0.24 to 0.46 | 61 | 0.31 | 97 | 0.7 |

Note. — KL, Kellgren and Lawrence

Original (from Grade 0 to Grade 4): No osteoarthritis; Doubtful narrowing of joint space and possible osteophytic lipping; Definite osteophytes and possible narrowing of joint space; Multiple osteophytes, definite narrowing of joint space and some sclerosis and possible deformity of bone ends; Large osteophyte, marked narrowing of joint space, severe sclerosis and definite deformity of bone ends.

Alternative 1 (from Grade 0 to Grade 4): No osteoarthritis; Possible osteophytes; Definite osteophytes; Osteophytes and JSN; Large osteophytes, marked narrowing of joint space and definite deformity.

Alternative 2 (from Grade 0 to Grade 4): No osteoarthritis; Minute osteophyte, doubtful significance; Definite osteophytes, unimpaired joint space; Moderate diminution of joint space (with osteophyte); Joint space greatly impaired with sclerosis of subchondral bone.

Alternative 3 (from Grade 0 to Grade 4): No osteoarthritis; Possible osteophytes only; Definite osteophytes and possible JSN; Moderate osteophytes and/or definite narrowing; Large osteophytes, severe JSN and/or bony sclerosis.

Alternative 4 (from Grade 0 to Grade 4): No osteoarthritis; Possible osteophytic lipping; Definite osteophytes and possible JSN; Moderate multiple osteophytes, definite JSN, some sclerosis and possible bone contour deformity (bony attrition); Large osteophytes, marked JSN, severe sclerosis and definite bony contour deformity (bony attrition).